%% file: naacl2021.tex
\pdfoutput=1

\documentclass[11pt]{article}

\usepackage{naacl2021}

\usepackage{times}
\usepackage{latexsym}

\usepackage[T1]{fontenc}

\usepackage[utf8]{inputenc}

\usepackage{microtype}
\usepackage[normalem]{ulem}
\usepackage{times}
\usepackage{latexsym}
\usepackage{url}
\usepackage{multirow}
\usepackage{balance}
\usepackage{subfigure}
\usepackage{latexsym}
\usepackage{amsmath}
\usepackage{amssymb}
\usepackage{caption}
\usepackage{graphicx}
\usepackage{url}
\usepackage{multicol}
\usepackage{multirow}
\usepackage{booktabs}
\usepackage{color}
\usepackage{subfigure}
\usepackage{array}
\usepackage{balance}
\usepackage{appendix}
\usepackage{amsmath}
\usepackage{graphicx} 
\title{Neural Quality Estimation with Multiple Hypotheses for Grammatical Error Correction}

\author{Zhenghao Liu$^{1,2}$, Xiaoyuan Yi$^{1,2}$, Maosong Sun$^{1,3}$\thanks{ \ \ Corresponding author: M. Sun (sms@tsinghua.edu.cn)}, Liner Yang$^{4}$, Tat-Seng Chua$^{5}$ \\ 
$^1$Department of Computer Science and Technology, Tsinghua University, Beijing, China\\
Institute for Artificial Intelligence, Tsinghua University, Beijing, China\\
Beijing National Research Center for Information Science and Technology\\
$^2$State Key Lab on Intelligent Technology and Systems, Tsinghua University, Beijing, China\\
$^3$Beijing Academy of Artificial Intelligence \\
$^4$Beijing Language and Culture University, Beijing, China \\
$^5$School of Computing, National University of Singapore, Singapore \\}

\begin{document}
\maketitle
\vspace{5em}
\input{abstract.tex}
\input{introduction.tex}
\input{relatedwork.tex}
\input{method.tex}

\input{experiment.tex}
\input{result.tex}
\input{conclusion.tex}
\section*{Acknowledgments}
We thank the reviewers and Shuo Wang for their valuable comments and advice.
This research is mainly supported by Science \& Tech Innovation 2030 Major Project “New Generation AI” (Grant no. 2020AAA0106500) as well as supported in part by a project from Shanghai-Tsinghua International Innovation Center and the funds of Beijing Advanced Innovation Center for Language Resources under Grant TYZ19005.

\nobalance

\bibliography{citation}
\bibliographystyle{acl_natbib}
\clearpage
\appendix
\input{appendix}

\end{document}

%% file: abstract.tex
\begin{abstract}
Grammatical Error Correction (GEC) aims to correct writing errors and help language learners improve their writing skills. However, existing GEC models tend to produce spurious corrections or fail to detect lots of errors. The quality estimation model is necessary to ensure learners get accurate GEC results and avoid misleading from poorly corrected sentences. Well-trained GEC models can generate several high-quality hypotheses through decoding, such as beam search, which provide valuable GEC evidence and can be used to evaluate GEC quality. However, existing models neglect the possible GEC evidence from different hypotheses. This paper presents the Neural Verification Network (VERNet) for GEC quality estimation with multiple hypotheses. VERNet establishes interactions among hypotheses with a reasoning graph and conducts two kinds of attention mechanisms to propagate GEC evidence to verify the quality of generated hypotheses. Our experiments on four GEC datasets show that VERNet achieves state-of-the-art grammatical error detection performance, achieves the best quality estimation results, and significantly improves GEC performance by reranking hypotheses. All data and source codes are available at \url{https://github.com/thunlp/VERNet}.
\end{abstract}

%% file: introduction.tex
\section{Introduction}
Grammatical Error Correction (GEC) systems primarily aim to serve second-language learners for proofreading. These systems are expected to detect grammatical errors, provide precise corrections, and guide learners to improve their language ability. With the rapid increase of second-language learners, GEC has drawn growing attention from numerous researchers of the NLP community.

Existing GEC systems usually inherit the seq2seq architecture~\cite{sutskever2014sequence} to correct grammatical errors or improve sentence fluency. These systems employ beam search decoding to generate correction hypotheses and rerank hypotheses with quality estimation models from $K$-best decoding~\cite{kiyono2019empirical,kaneko2020encoder} or model ensemble~\cite{chollampatt2018multilayer} to produce more appropriate and accurate grammatical error corrections. Such models thrive from edit distance and language models~\cite{chollampatt2018multilayer,chollampatt2019cross,yannakoudakis2017neural,kaneko2019tmu,kaneko2020encoder}.
\citet{chollampatt2018neural} further consider the GEC accuracy in quality estimation by directly predicting the official evaluation metric, F$_{0.5}$ score.
\input{figure/ppl}

The $K$-best hypotheses from beam search usually derive from model uncertainty~\cite{ott2018analyzing}. These uncertainties of multi-hypotheses come from model confidence and potential ambiguity of linguistic variation~\cite{fomicheva2020multi}, which can be used to improve machine translation performance~\cite{wang2019improving}. \citet{fomicheva2020multi} further leverage multi-hypotheses to make convinced machine translation evaluation, which is more correlated with human judgments. 
Their work further demonstrates that multi-hypotheses from well-trained neural models have the ability to provide more hints to estimate generation quality. 
\input{figure/pr_beam}

For GEC, the hypotheses from the beam search decoding of well-trained GEC models can provide some valuable GEC evidence. We illustrate the reasons as follows.
\begin{itemize}
    \item \textit{Beam search can provide better GEC results.} The GEC performance of the top-ranked hypothesis and the best one has a large gap in beam search. For two existing GEC systems, \citet{zhao2019improving} and \citet{kiyono2019empirical}, the F$_{0.5}$ scores of these systems are 58.99 and 62.03 on the CoNLL2014 dataset. However, the F$_{0.5}$ scores of the best GEC results of these systems can achieve 73.56 and 76.82.
    \item \textit{Beam search candidates are more grammatical.} As shown in Figure~\ref{fig:ppl}, the hypotheses from well-trained GEC models with beam search usually win the favor of language models, even for these hypotheses ranked to the rear. It illustrates these hypotheses are usually more grammatical than source sentences.
    \item \textit{Beam search candidates can provide valuable GEC evidence.} As shown in Figure~\ref{fig:beam_pr}, the hypotheses of different beam ranks have almost the same Recall score, which demonstrates all hypotheses in beam search can provide some valuable GEC evidence.
\end{itemize}

Existing quality estimation models~\cite{chollampatt2018neural} for GEC regard hypotheses independently and neglect the potential GEC evidence from different hypotheses.
To fully use the valuable GEC evidence from GEC hypotheses, we propose the Neural Verification Network (VERNet) to estimate the GEC quality with modeled interactions from multi-hypotheses.
Given a source sentence and $K$ hypothesis sentences from the beam search decoding of the basic GEC model, VERNet establishes hypothesis interactions by regarding $\langle$source, hypothesis$\rangle$ pairs as nodes, and constructing a fully-connected reasoning graph to propagate GEC evidence among multi-hypotheses. Then VERNet proposes two kinds of attention mechanisms on the reasoning graph, \emph{node interaction attention} and \emph{node selection attention}, to summarize and aggregate necessary GEC evidence from other hypotheses to estimate the quality of tokens.

Our experiments show that VERNet can pick up necessary GEC evidence from multi-hypotheses provided by GEC models and help verify the quality of GEC hypotheses. VERNet helps GEC models to generate more accurate GEC results and benefits most grammatical error types.

%% file: figure/ppl.tex
\begin{figure}[t]
    \centering
    \subfigure[BEA19.] { \label{fig:bea19ppl} 
    \includegraphics[width=0.45\linewidth]{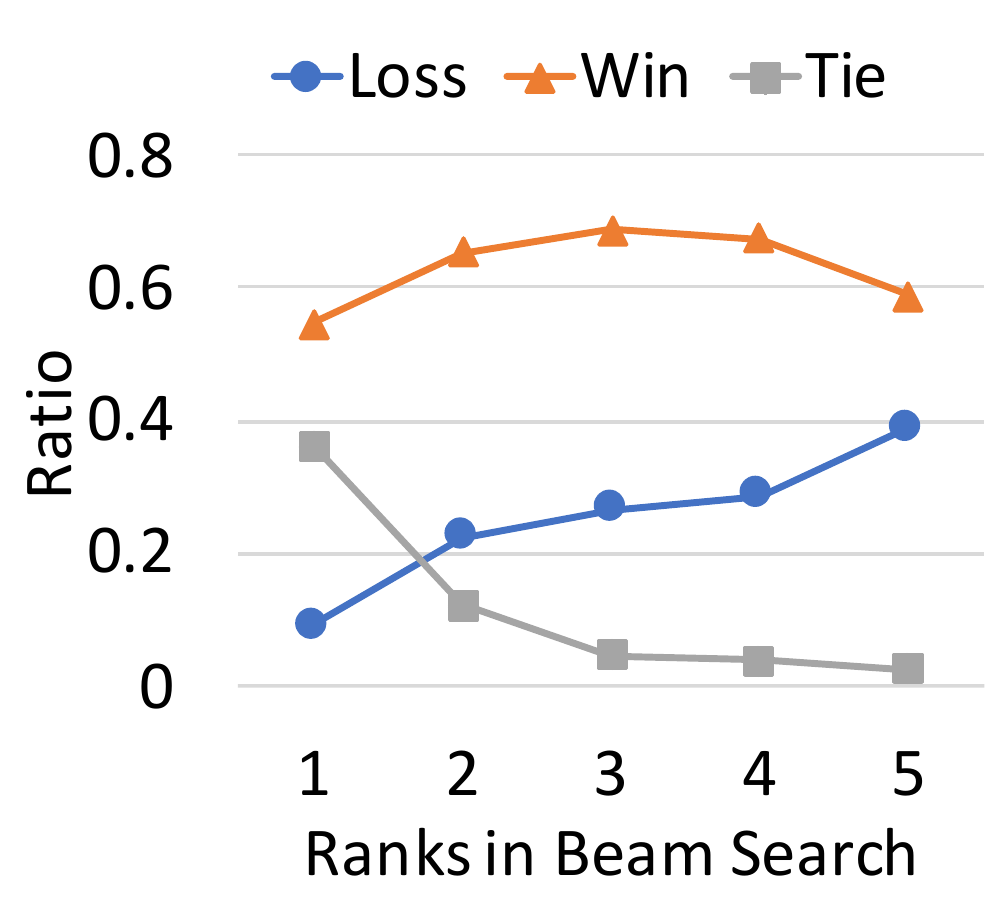}}
    \subfigure[CoNLL2014.] { \label{fig:conll14ppl} 
    \includegraphics[width=0.45\linewidth]{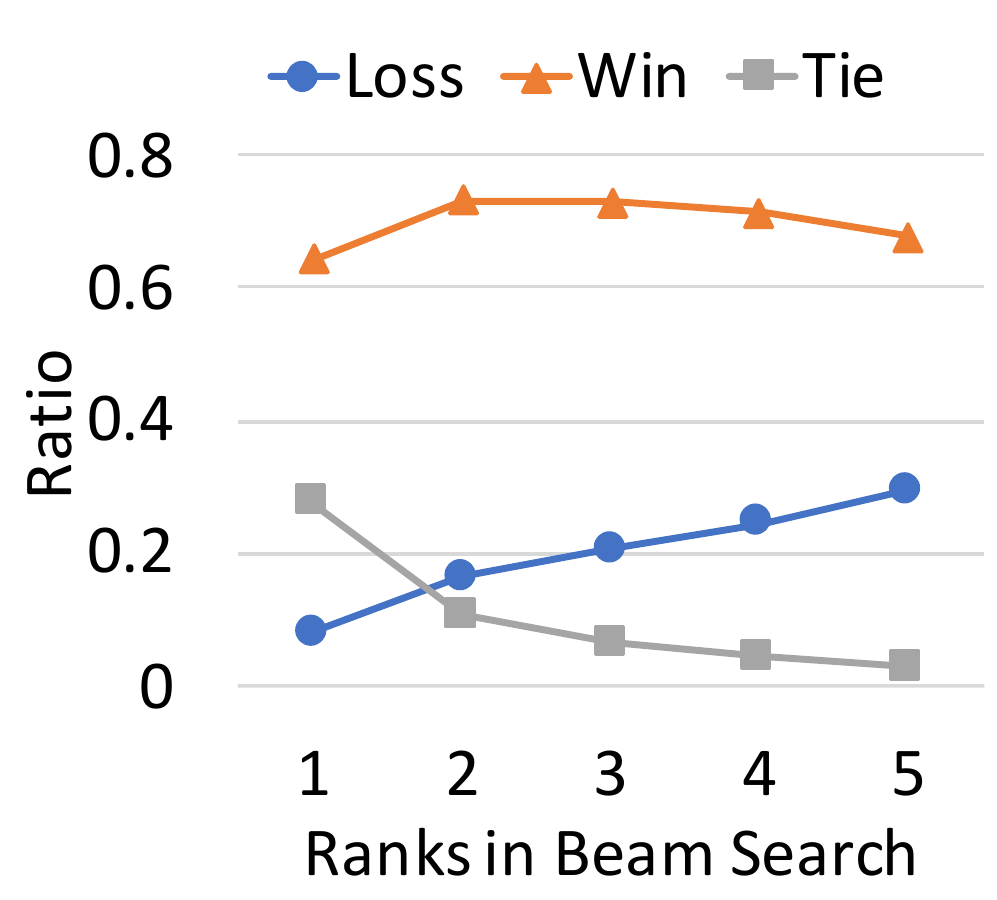}}
    \caption{The Grammaticality of Generated Hypotheses. The hypotheses are generated by~\citet{kiyono2019empirical} with beam search decoding. The hypothesis is compared to the source sentence with a BERT based language model and classified into Win (the hypothesis is better), Tie (the hypothesis and source are same) and Loss (the source is better). The ratios of different classes are plotted with different beam search ranks.}
    \label{fig:ppl}
\end{figure}

%% file: figure/pr_beam.tex
\begin{figure}[t]
    \centering
    \subfigure[CoNLL2014 (ann. 1).] { \label{fig:conll14:ann0} 
    \includegraphics[width=0.45\linewidth]{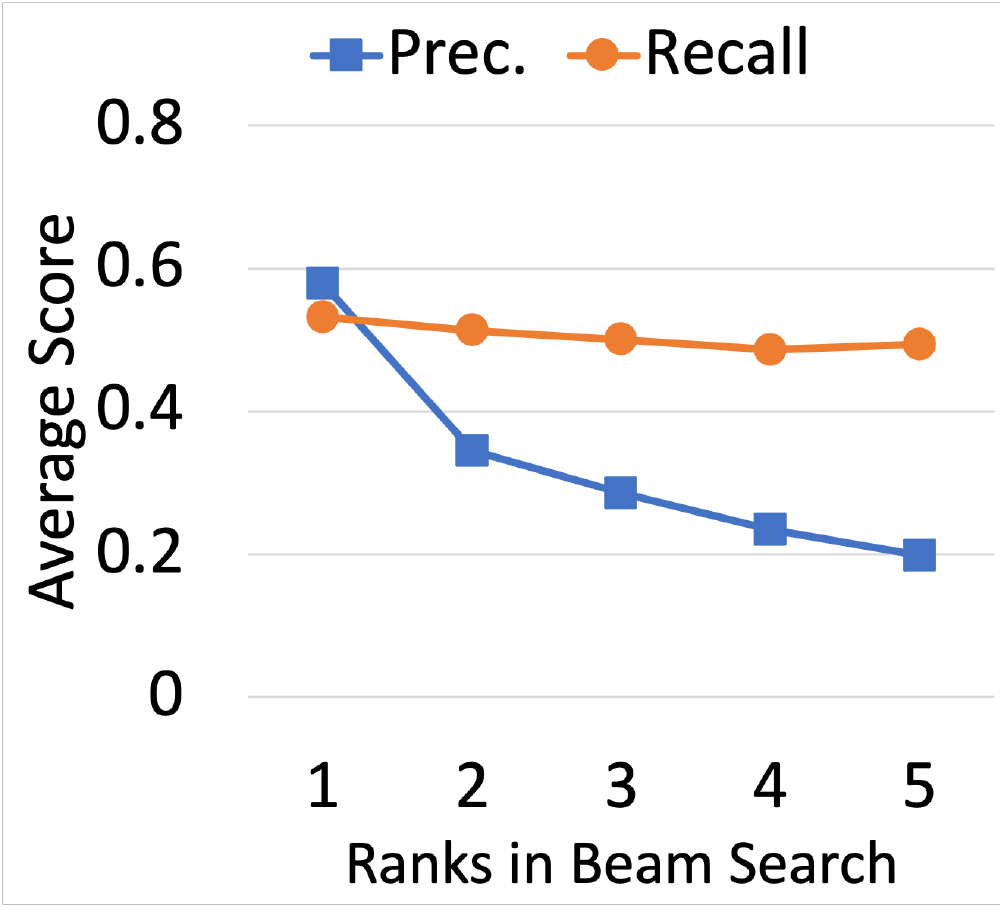}}
    \subfigure[CoNLL2014 (ann. 2).] { \label{fig:conll14:ann1} 
    \includegraphics[width=0.45\linewidth]{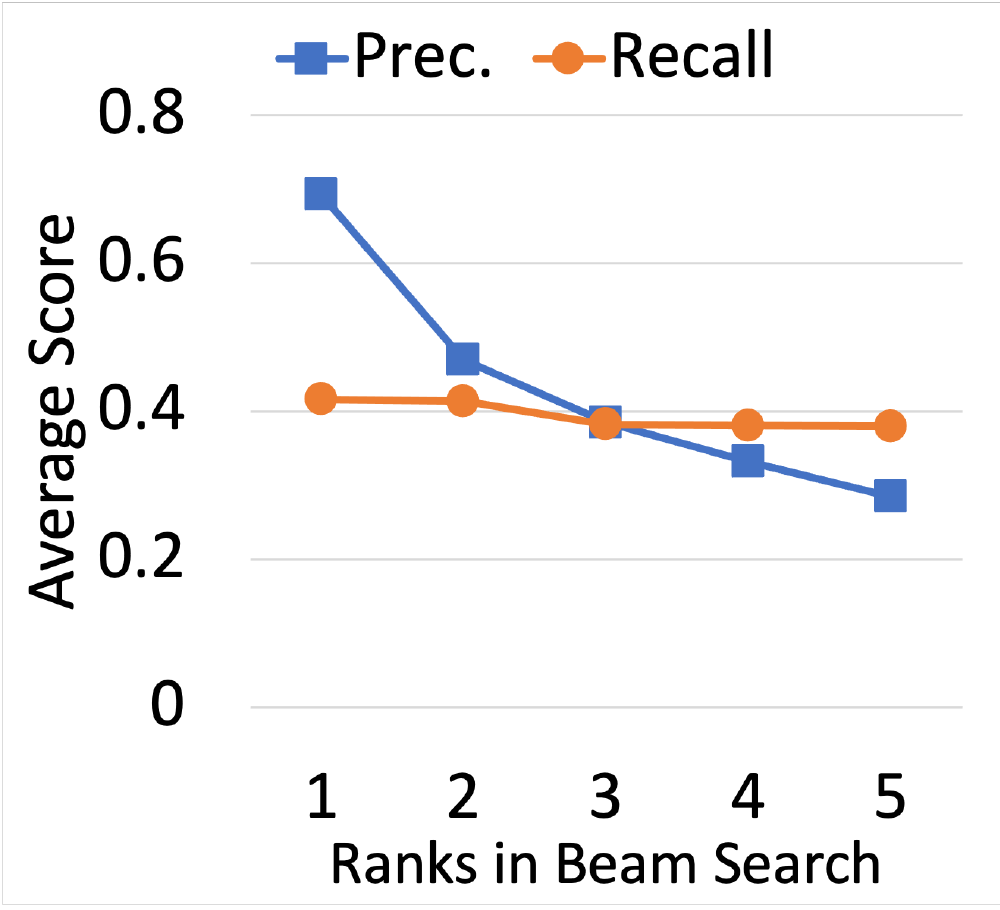}}
    \caption{The GEC Performance of Generated Hypotheses. The hypotheses generated  by~\citet{kiyono2019empirical} are evaluated on the CoNLL2014 dataset. The average scores of Precision and Recall are calculated according to the two annotations of CoNLL2014.}
    \label{fig:beam_pr}
\end{figure}

%% file: relatedwork.tex
\section{Related Work}
The GEC task is designed for automatically proofreading. Large-scale annotated corpora~\cite{Mizumoto2011MiningRL,dahlmeier2013building,bryant2019bea} bring an opportunity for building fully data-driven GEC systems.

Existing neural models regard GEC as a natural language generation (NLG) task and usually use sequence-to-sequence architecture~\cite{sutskever2014sequence} to generate correction hypotheses with beam search decoding~\cite{yuan2016grammatical,chollampatt2018multilayer}. Transformer-based architectures~\cite{Vaswani2017AttentionIA} show their effectiveness in NLG tasks and are also employed to achieve convinced correction results~\cite{grundkiewicz2019neural,kiyono2019empirical}. The copying mechanism is also introduced for GEC models~\cite{zhao2019improving} to better align tokens from source sentence to hypothesis sentence. To further accelerate the generation process, some work also comes up with non-autoregressive GEC models and leverages a single encoder to parallelly detect and correct grammatical errors~\cite{awasthi2019parallel,malmi2019encode,omelianchuk2020gector}.

Recent research focuses on two directions to improve GEC systems. The first one treats GEC as a low-resource language generation problem and focuses on data augmentation for a grammar sensitive and language proficient GEC system~\cite{junczys2018approaching,kiyono2019empirical}. Various weak-supervision corpora have been leveraged, such as Wikipedia edit history~\cite{lichtarge2019corpora}, Github edit history~\cite{hagiwara2019github} and confusing word set~\cite{grundkiewicz2019neural}. Besides, lots of work generates grammatical errors through generation models or round-trip translation~\cite{ge2018fluency,wang2019denoising,xie2018noising}. \citet{kiyono2019empirical} further consider different data augmentation strategies to conduct better GEC pretraining. 

Reranking GEC hypotheses from $K$-best decoding or GEC model ensemble~\cite{hoang2016exploiting,chollampatt2018neural} with quality estimation models provides another promising direction to achieve better GEC performance. Some methods evaluate if hypotheses satisfy linguistic and grammatical rules. For this purpose, they employ language models~\cite{chollampatt2018multilayer,chollampatt2019cross} or grammatical error detection (GED) models to estimate hypothesis quality. GED models~\cite{rei2017semi,rei2019jointly} estimate the hypothesis quality 
on both sentence level~\cite{kaneko2019tmu} and token level~\cite{yannakoudakis2017neural}.
\citet{chollampatt2018neural} further estimate GEC quality by considering correction accuracy. They establish source-hypothesis interactions with the encoder-decoder architecture and learn to directly predict the official evaluation score F$_{0.5}$. 

The pre-trained language model BERT~\cite{devlin2019bert} has proven its effectiveness in producing contextual token representations, achieving better quality estimation~\cite{kaneko2019tmu,chollampatt2019cross} and improving GEC performance by fuse BERT representations~\cite{kaneko2020encoder}. However, existing quality estimation models regard each hypothesis independently and neglect the interactions among multi-hypotheses, which can also benefit the quality estimation~\cite{fomicheva2020multi}.

%% file: method.tex
\section{Neural Verification Network}
\input{figure/model}
This section describes Neural Verification Network (VERNet) to estimate the GEC quality with multi-hypotheses, as shown in Figure~\ref{fig:model}.

Given a source sentence $s$ and $K$ corresponding hypotheses $C = \{c^1, \dots, c^k, \dots, c^K\}$ generated by a GEC model, we first regard each source-hypothesis pair $\!\langle\!s, c^k\rangle$ as a node and fully connect all nodes to establish multi-hypothesis interactions. Then VERNet leverages BERT to get the representation of each token in $\!\langle\!s, c^k\rangle$ pairs (Sec.~\ref{model:bert}) and conducts two kinds of attention mechanisms to propagate and aggregate GEC evidence from other hypotheses to verify the token quality (Sec.~\ref{model:attention}). Finally, VERNet estimates hypothesis quality by aggregating token level quality estimation scores (Sec.~\ref{model:estimation}). Our VERNet is trained end-to-end with supervisions from golden labels (Sec.~\ref{model:training}).

\subsection{Initial Representations for Sentence Pairs}\label{model:bert}
Pre-trained language models, \textit{e.g.} BERT~\cite{devlin2019bert}, show their advantages of producing contextual token representations for various NLP tasks.
Hence, given a source sentence $s$ with $m$ tokens and the $k$-th hypothesis $c^k$ with $n$ tokens, we use BERT to encode the source-hypothesis pair $\!\langle\!s, c^k\!\rangle\!$ and get its representation $H^k$:
\begin{equation}
\small
    H^k = \text{BERT} (\text{[CLS]} \  s \  \text{[SEP]} \ c^k \ \text{[SEP]}).
\end{equation}
The pair representation $H^k$ consists of token-level representations, that is, $H^k= \{H_0^k, \dots, H_{m+n+2}^k\}$. $H_0^k$ denotes the representation of ``[CLS]'' token.

\subsection{Verify Token Quality with Multi-hypotheses}\label{model:attention}
VERNet conducts two kinds of attention mechanisms, \emph{node interaction attention} and \emph{node selection attention}, to verify the token quality with the verification representation $V^k$ of $k$-th node, which learns the supporting evidence towards estimating token quality from multi-hypotheses.

The node interaction attention first summarizes useful GEC evidence from the $l$-th node for the fine-grained representation $V^{l \rightarrow k}$ (Sec.~\ref{model:interaction_atten}). Then node selection attention further aggregates fine-grained representation $V^{l \rightarrow k}$ with score $\gamma^l$ according to each node's confidence (Sec.~\ref{model:node_atten}). Finally, we can calculate the verification representation $V^k$ to verify the token's quality of each node.

\subsubsection{Fine-grained Node Representation with Node Interaction Attention}\label{model:interaction_atten}
The node interaction attention $\alpha^{l \rightarrow k}$ attentively reads tokens in the $l$-th node and picks up supporting evidence towards the $k$-th node to build fine-grained node representations $V^{l \rightarrow k}$.

For the $p$-th token in the $k$-th node, $w_p^k$ , we first calculate the node interaction attention weight $\alpha_q^{l \rightarrow k}$ according to the relevance between $w_p^k$ and the $q$-th token in the $l$-th node, $w_q^l$:
\begin{equation}
\small
\alpha_q^{l \rightarrow k} = \text{softmax}_q  ((H^k_{p})^T \cdot W \cdot H^l_{q}),
\end{equation} 
where $W$ is a parameter. $H^k_{p}$ and $H^l_{q}$ are the representations of $w_p^k$ and $w_q^l$. Then all token representations of $l$-th node are aggregated:
\begin{equation}
\small
V^{l \rightarrow k}_p = \sum_{q=1}^{m+n+2} (\alpha_q^{l \rightarrow k} \cdot H^l_{q}).
\end{equation}

Based on $V^{l \rightarrow k}_p$, we further build the $l$-th node fine-grained representation towards the $k$-th node, $V^{l \rightarrow k} = \{V^{l \rightarrow k}_1,\dots, V^{l \rightarrow k}_p, \dots , V^{l \rightarrow k}_{m+n+2}\}$.

\subsubsection{Evidence Aggregation with Node Selection Attention}\label{model:node_atten}
The node selection attention measures node importance and is used to aggregate supporting evidence from the fine-grained node representation $V^{l \rightarrow k}$ of the $l$-th node. We leverage attention-over-attention mechanism~\cite{cui2017attention} to conduct source $h^{ls}$ and hypotheses $h^{lh}$ representations to calculate the $l$-th node selection attention score $\gamma^l$. Then we get the node verification representation $V_p^k$ with the node selection attention $\gamma^l$.

To calculate the node selection attention $\gamma^l$, we establish an interaction matrix $M^l$ between the source and hypothesis sentences of the $l$-th node. Each element $M^l_{ij}$ in $M^l$ is calculated with the relevance between $i$-th source token and $j$-th hypothesis token (include ``[SEP]'' tokens):
\begin{equation}
\small
M^l_{ij} = (H^l_{i})^T \cdot W \cdot H^l_{m + 1 + j},
\end{equation}
where $W$ is a parameter. Then we calculate attention scores $\beta_{i}^{ls}$ and $\beta_{j}^{lh}$ along the source dimension and hypothesis dimension, respectively:

\begin{small}
\begin{align}
\beta_{i}^{ls} &= \frac{1}{n+1}\sum_{j=1}^{n+1} \text{softmax}_i(M^l_{ij}),\\
\beta_{j}^{lh} &= \frac{1}{m+1}\sum_{i=1}^{m+1} \text{softmax}_j(M^l_{ij}).
\end{align}
\end{small}

Then the representations of source sentence and hypothesis sentence are calculated: 
\begin{small}
\begin{equation}
h^{ls} = \sum_{i=1}^{m+1} \beta_{i}^{ls} \cdot H_i^l, \ \ \
h^{lh} = \sum_{j=1}^{n+1} \beta_{j}^{lh} \cdot H_{m + 1 + j}^l.
\end{equation}
\end{small}

Finally, the node selection attention $\gamma^l$ of $l$-th node is calculated for the evidence aggregation:
\begin{equation}
\small
\gamma^l = \text{softmax}_l(\text{Linear}((h^{ls} \circ h^{lh}) ; h^{ls} ; h^{lh})),
\end{equation}
where $\circ$ is the element-wise multiplication operator and $;$ is the concatenate operator.

The node selection attention $\gamma^l$ aggregates evidence for the verification representation $V_p^k$ of $w_p^k$:
\begin{equation}
\small
V_p^k = \sum_{l=1}^{K} (\gamma^l \cdot V^{l \rightarrow k}_p),
\end{equation}
where $V^{k} = \{V^{k}_1,\dots, V^{k}_p, \dots , V^{k}_{m+n+2}\}$ is the $k$-th node verification representation.

\subsection{Hypothesis Quality Estimation}\label{model:estimation}
For the $p$-th token $w_p^k$ in the $k$-th node, the probability $P(y|w_p^k)$ of quality label $y$ is calculated with the verification representation $V_p^k$:
\begin{equation}\label{eq:prob}
\small
P(y|w_p^k) = \text{softmax}_y (\text{Linear}((H^k_p \circ V_p^k) ; H^k_p ; V_p^k)),
\end{equation}
where $\circ$ is the element-wise multiplication and $;$ is the concatenate operator. We average all probability $P(y=1|w_p^k)$ of token level quality estimation as hypothesis quality estimation score $f(s, c^k)$ for the pair $\langle s, c^k\rangle$:
\begin{equation}
\small
    f(s, c^k) = \frac{1}{n+1} \sum_{p=m+2}^{m+n+2} P(y=1|w_p^k).
\end{equation}

\subsection{End-to-end Training}\label{model:training}
We conduct joint training with token-level supervision. The source labels and hypothesis labels are used, which denote the grammatical quality of source sentences and GEC accuracy of hypotheses.

The cross entropy loss for the $p$-th token $w_p^k$ in the $k$-th node is calculated:
\begin{equation}
\small
L(w_p^k) = \text{CrossEntropy}(y^*, P(y|w_p^k)),
\end{equation}
using the ground truth token labels $y^*$.

Then the training loss of VERNet is calculated:
\begin{equation}
\small
L = \frac{1}{K} \frac{1}{m+n+2} \sum_{k=1}^K \sum_{p=1}^{m+n+2} L(w_p^k).
\end{equation}

%% file: figure/model.tex
\begin{figure}[t]
    \centering
    \includegraphics[width=\linewidth]{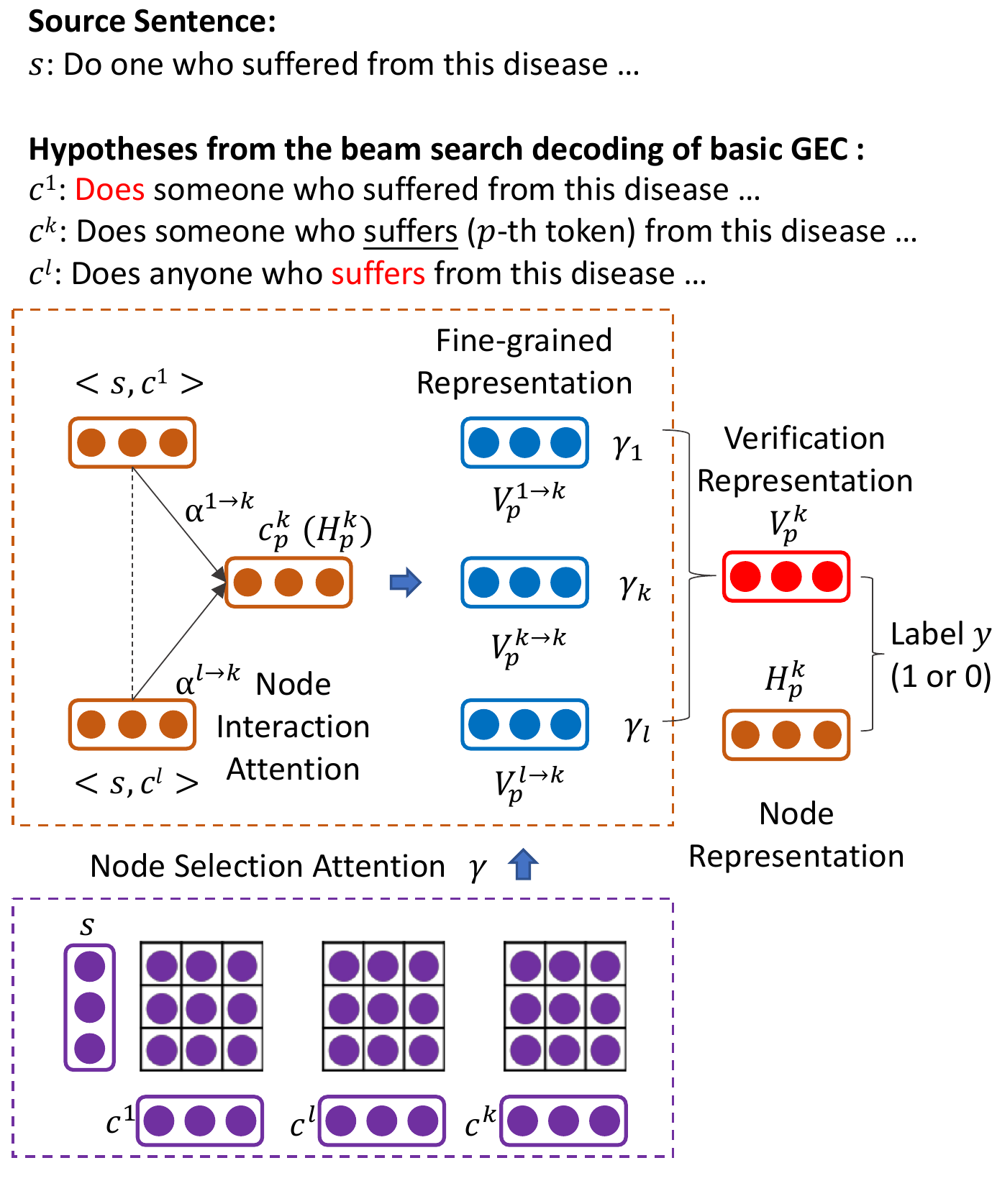}
    \caption{The Architecture of Neural Verification Network (VERNet). The \underline{estimated token} ($c_p^k$) and potentially \textbf{\textcolor{red}{supporting evidence}} towards $c_p^k$ are annotated.}
    \label{fig:model}
\end{figure}

%% file: experiment.tex
\section{Experimental Methodology}
This section describes the datasets, evaluation metrics, baselines, and implementation details.

\textbf{Datasets.}
We use FCE~\cite{yannakoudakis2011new}, BEA19~\cite{bryant2019bea} and NUCLE~\cite{dahlmeier2013building} to construct training and development sets. Four testing scenarios, FCE, BEA19 (Restrict), CoNLL-2014~\cite{ng2014conll} and JFLEG~\cite{napoles2017jfleg}, are leveraged to evaluate model performance. Detailed data statistics are presented in Table~\ref{tab:dataset}. We do not incorporate additional training corpora for fair comparison.

\textbf{Basic GEC Model}.
To generate correction hypotheses, we take one of the state-of-the-art autoregressive GEC systems~\cite{kiyono2019empirical} as our \emph{basic GEC model} and keep the same setting. The beam size of our baseline model is set to 5~\cite{kiyono2019empirical}, and all these beam search hypotheses are reserved in our experiments.

We generate quality estimation labels for tokens in both source sentences and hypothesis sentences with ERRANT~\cite{bryant2017automatic,felice2016automatic}, which indicate grammatical correctness and GEC accuracy, respectively. As shown in Table~\ref{tab:annotation_exp}, ERRANT annotates edit operations (delete, insert, and replace) towards the ground truth corrections. In terms of such annotations, each token is labeled with correct (1) or incorrect (0).

\textbf{Evaluation Metrics.} We introduce the evaluation metrics in three tasks: token quality estimation, sentence quality estimation, and GEC.  

To evaluate the model performance of token-level quality estimation, we employ the same evaluation metrics from previous GED models~\cite{rei2017semi,rei2019jointly,yannakoudakis2017neural}, including Precision, Recall, and F$_{0.5}$. F$_{0.5}$ is our primary evaluation metric. 
\input{table/data.tex}
\input{table/example_ann.tex}

For the evaluation of sentence-level quality estimation, we employ the same evaluation metrics from the previous quality estimation model~\cite{chollampatt2018neural}, including two evaluation scenarios: (1) GEC evaluation metrics for the hypothesis that reranked top-1 and (2) Pearson Correlation Coefficient (PCC) between reranking scores and golden scores (F$_{0.5}$) for all hypotheses. 

To evaluate GEC performance, we adopt GLEU~\cite{napoles2015ground} to evaluate model performance on the JFLEG dataset. The official tool ERRANT of the BEA19 shared task~\cite{bryant2019bea} is used to calculate Precision, Recall, and F$_{0.5}$ scores for other datasets. For the CoNLL-2014 dataset, the M$^2$ evaluation~\cite{dahlmeier2012better} is also adopted as our main evaluation.

\textbf{Baselines.} BERT-fuse (GED)~\cite{kaneko2020encoder} is compared in our experiments, which trains BERT with the GED task and fuses BERT representations into the Transformer.
For quality estimation, we consider two groups of baseline models in our experiments, and more details of these models can be found in Appendices~\ref{appendix:sentence_score}.

(1) \emph{BERT based language models.} We employ three BERT based language models to estimate the quality of hypotheses. BERT-LM~\cite{chollampatt2019cross} measures hypothesis quality with the perplexity of the language model. BERT-GQE~\cite{kaneko2019tmu} is trained with annotated GEC data and estimates if the hypothesis has grammatical errors. We also conduct BERT-GED (SRC) that predicts token level grammar indicator labels, which is inspired by GED models~\cite{yannakoudakis2017neural}. BERT shows significant improvement compared to LSTM based models for the GED task (Appendices~\ref{appendix:ged}). Hence the LSTM based models are neglected in our experiments.

(2) \emph{GEC accuracy estimation models.} These models further consider the source-hypothesis interactions to evaluate GEC accuracy. We take a strong baseline NQE~\cite{chollampatt2018neural} in experiments. NQE employs the encoder-decoder (predictor) architecture to encode source-hypothesis pairs and predicts F$_{0.5}$ score with the estimator architecture.  All their proposed architectures, NQE (CC), NQE (RC), NQE (CR), and NQE (RR) are compared. For NQE (XY), X indicates the predictor architecture, and Y indicates the estimator architecture. X and Y can be recurrent (R) or convolutional (C) neural networks. In addition, we also employ BERT to encode source-hypothesis pairs and then predict the F$_{0.5}$ score to implement the BERT-QE model.
We also come up with two baselines, BERT-GED (HYP) and BERT-GED (JOINT). They leverage BERT to encode source-hypothesis pairs and are supervised with the token-level quality estimation label. BERT-GED (HYP) is trained with the supervision of hypotheses, and BERT-GED (JOINT) is supervised with labels from both source and hypothesis sentences.

\input{table/detection1}
\input{table/feature}

\textbf{Implementation Details.}
In all experiments, we use the base version of BERT~\cite{devlin2019bert} and ELECTRA~\cite{clark2019electra}. BERT is a widely used pretrained language model and trained with the mask language model task. ELECTRA is trained with the replaced token detection task and aims to predict if the token is original or replaced by a BERT based generator during pretraining. ELECTRA is a discriminator based pretrained language model and is more like the GED task. We regard BERT as our main model for text encoding and leverage ELECTRA to evaluate the generalization ability of our model.

Both BERT and ELECTRA inherit huggingface's PyTorch implementation~\cite{wolf2020transformers}. Adam~\cite{kingma2014adam} is utilized for parameter optimization. We set the max sentence length to 120 for source and hypothesis sentences, learning rate to 5e-5, batch size to 8, and accumulate step to 4 during training.

For hypothesis reranking, we leverage the learning-to-rank method, Coordinate Ascent (CA)~\cite{metzler2007linear}, to aggregate the ranking features and basic GEC score to conduct the ranking score. We assign the hypotheses with the highest F$_{0.5}$ score as positive instances and the others as negative ones. The Coordinate Ascent method is implemented by RankLib\footnote{\url{https://sourceforge.net/p/lemur/wiki/RankLib/}}.

%% file: table/data.tex
\begin{table}[t]
\begin{center}
\small
\begin{tabular}{l | c c c}
\hline 
Dataset & Training & Development & Test  \\ \hline
FCE & 28,350 & 2,191 & 2,695 \\
BEA19 & 34,308 & 4,384 & 4,477 \\
NUCLE & 57,151 & - & - \\
CoNLL-2014 & - & - & 1,312 \\ 
JFLEG & - & - & 747 \\\hline
Total & 119,809 & 6,575 & 9,231 \\
\hline
\end{tabular}
\end{center}
\caption{\label{tab:dataset}Data Statistics.}
\end{table}

%% file: table/example_ann.tex
\begin{table}[t]
\begin{center}
\small
\begin{tabular}{ l | p{1.5cm}<{\centering}  p{1.5cm}<{\centering}  p{1.5cm}<{\centering}}
\hline
Sentence & \multicolumn{3}{c}{\begin{tabular}[c]{@{}l@{}}The $_1$ \textcolor{red}{a} $_2$ Mobile phone is a marvelous\\ invention to $_9$ \textcolor{red}{charge} $_{10}$ the world $_{12}$ \textcolor{red}{[SEP]}\end{tabular}} \\ \hline
\multirow{4}{*}{Correction} & Operation & Span & Edit \\ \cline{2-4}
&Delete & 1,2 & - \\
& Replace & 9,10 & change \\
& Insert & 12,12 & . \\
\hline
\end{tabular}
\end{center}
\caption{\label{tab:annotation_exp}An Example of Token Label Annotation. All sentences are annotated with ERRANT according to the golden correction. The \textcolor{red}{words} in red color are labeled as incorrect (0) and others are labeled as correct (1). The ``[SEP]'' token denotes the end of the sentence.}
\end{table}

%% file: table/detection1.tex
\begin{table*}[t]
\begin{center}
\small
\begin{tabular}{l | l | p{0.7cm}<{\centering} p{0.7cm}<{\centering} p{0.7cm}<{\centering} | p{0.7cm}<{\centering} p{0.7cm}<{\centering} p{0.7cm}<{\centering} | p{0.7cm}<{\centering} p{0.7cm}<{\centering} p{0.7cm}<{\centering}}
\hline
& \multirow{2}{*}{Model} & \multicolumn{3}{c}{FCE test set} & \multicolumn{3}{|c}{CoNLL-2014 ann. 1} & \multicolumn{3}{|c}{CoNLL-2014 ann. 2} \\ \cline{3-11}
& & P & R & F$_{0.5}$ & P & R & F$_{0.5}$ & P & R & F$_{0.5}$\\
\hline
\multirow{4}{*}{Source} & BERT-GED (SRC) & 74.22 & 43.34 & 64.97 & 59.84 & 27.11 & 48.20 & 77.94 & 25.02 & 54.77 \\
&BERT-GED (JOINT) & 75.62 & 44.44 & 66.32 & 60.79 & 27.33 & 48.83 & 77.42 & 25.23 & 54.77 \\
&BERT-VERNet & \textbf{81.53} & 45.71 & 70.48 & \textbf{62.64} & 30.62 & 51.80 & \textbf{82.25} & 28.49 & 59.71 \\
&ELECTRA-VERNet & 80.94 & \textbf{50.51} & \textbf{72.24} & 62.50 & \textbf{35.61} & \textbf{54.30} & 81.69 & \textbf{32.97} & \textbf{63.06} \\
\hline
\multirow{4}{*}{Hypothesis} & BERT-GED (HYP) & 80.27 & 40.58 & 67.14 & 74.28 & 34.20 & 60.17 & 66.49 & 27.68 & 51.93 \\
&BERT-GED (JOINT) & 76.71 & 46.94 & 68.07 & 71.15 & 38.30 & 60.73 & 64.79 & 31.52 & 53.50 \\
&BERT-VERNet & \textbf{81.85} & 44.27 & 69.97 & \textbf{76.03} & 34.02 & 60.97 & 71.79 & 29.04 & 55.46\\
&ELECTRA-VERNet & 80.62 & \textbf{49.16} & \textbf{71.48} & 74.80 & \textbf{39.26} & \textbf{63.33} & \textbf{72.55} & \textbf{34.42} & \textbf{59.39} \\
\hline
\end{tabular}
\caption{\label{tab:lab_acc} Performance of Token Level Quality Estimation. Both source and hypothesis scenarios are conducted to evaluate grammatical quality estimation ability on source sentences and GEC quality estimation ability on hypotheses, respectively. BERT-GED (SRC) only encodes source sentences while others encode $\!\langle \!$ source, hypothesis $\!\rangle\!$ pairs. BERT-GED (JOINT) is supervised with golden labels from source and hypothesis sentences.
}
\end{center}
\end{table*}

%% file: table/feature.tex
\begin{table*}
\small
\resizebox{0.99\textwidth}{!}{
\begin{tabular}{l|  c  c  c  c 
 c|  c  c  c  c |  c  c  c |
 c  c}
\hline \multirow{3}{*}{Model} & \multicolumn{5}{c|}{CoNLL-2014 (M$^2$)} &  \multicolumn{4}{c|}{FCE} &
\multicolumn{3}{c|}{BEA19} & \multicolumn{2}{c}{JFLEG}\\ \cline{2-15}
& \multirow{2}{*}{P} & \multirow{2}{*}{R} & \multirow{2}{*}{F$_{0.5}$} & PCC & PCC  & \multirow{2}{*}{P} & \multirow{2}{*}{R} & \multirow{2}{*}{F$_{0.5}$} & \multirow{2}{*}{PCC} & \multirow{2}{*}{P} & \multirow{2}{*}{R} & \multirow{2}{*}{F$_{0.5}$}
&\multirow{2}{*}{GLEU} & \multirow{2}{*}{PCC}\\ 
&  &  &  &   (ann.1)  & (ann.2) & &  & &  &  &  & \\ \hline

NQE (RR) & 61.38 & 33.03 & 52.39 & 23.43 & 6.62 & 51.43 & 30.36 & 45.16 & 28.74 & 57.22 & 46.33 & 54.65 & 55.90 & 1.29 \\ 
NQE (RC) & 60.09 & 33.11 & 51.67 & 24.12 & 5.52 & 53.97 & 31.35 & 47.17 & 31.20 & 57.87 & 47.24 & 55.37 & 56.91 & 1.66 \\ 
NQE (CR) & 62.52 & 35.24 & 54.14 & 24.80 & 9.12 & 51.77 & 31.46 & 45.85 & 30.69 & 57.92 & 47.43 & 55.47 & 56.92 & 6.48 \\ 
NQE (CC) & 60.62 & 35.77 & 53.23 & 22.94 & 8.39 & 50.21 & 32.09 & 45.11 & 29.23 & 56.83 & 49.47 & 55.19 & 57.22 & 7.68 \\
\hline
BERT-LM & 52.82 & 49.59 & 52.14 & 3.47 & 17.62 & 36.97 & 43.42 & 38.10 & 8.59 & 46.32 & 64.05 & 49.03 & 59.72 & 26.85 \\
BERT-GQE & 52.67 & 50.39 & 52.19 & 2.56 & 14.54 & 36.05 & 43.53 & 37.33 & 10.18 & 46.15 & 64.01 & 48.88 & 60.17 & 29.05 \\
BERT-GED (SRC) & 52.98 & \textbf{52.07} & 52.79 & 3.78 & 20.56 & 37.58 & \textbf{45.81} & 38.98 & 12.71 & 47.15 & \textbf{65.09} & 49.90 & 60.32 & 27.28\\
\hline
BERT-QE& 62.24 & 38.27 & 55.31 & 22.85 & 12.17 & 52.01 & 36.89 & 48.07 & 33.84 & 58.63 & 54.19 & 57.69 & 59.73 & 26.16\\
BERT-GED (HYP)& 68.90 & 34.35 & 57.36 & 30.06 & 16.79 & 57.21 & 36.03 & 51.19 & 43.48 & 68.18 & 53.85 & 64.73 & 60.00 & 29.90\\
BERT-GED (JOINT)& 69.33 & 36.02 & 58.51 & 28.62 & 16.28 & 58.53 & 37.24 & 52.53 & 45.08 & 66.80 & 55.09 & 64.07 & 60.49 & 33.03 \\
\hline
BERT-VERNet & 68.75 & 40.26 & 60.22 & 31.02 & 22.75 & 58.32 & 39.99 & 53.42 & 47.19 & 66.86 & 58.60 & 65.02 & 61.36 & 36.98 \\
ELECTRA-VERNet & \textbf{69.97} & 42.12 & \textbf{61.80} & \textbf{37.18} & \textbf{28.77} & \textbf{58.77} & 41.86 & \textbf{54.37} & \textbf{48.12} & \textbf{69.09} & 60.91 & \textbf{67.28} & \textbf{61.61} & \textbf{38.63} \\
\hline
\end{tabular}}
\caption{\label{tab:feature_rank}Performance of Sentence Level Quality Estimation. The ranked top-1 hypothesis is used to calculate GEC metrics. NQE~\cite{chollampatt2018neural} uses RNN or CNN models for GEC quality estimation. BERT-LM~\cite{chollampatt2019cross} measures perplexity without fine-tuning. BERT-GQE~\cite{kaneko2019tmu} and BERT-GED (SRC) are supervised with sentence-level and token-level labels from source sentences to estimate grammatical quality, respectively. NQE and BERT-QE
encode $\!\langle \!$ source, hypothesis $\!\rangle\!$ pairs and directly predict F$_{0.5}$ score. BERT-GED (HYP) and BERT-GED (JOINT) encode the $\!\langle \!$ source, hypothesis $\!\rangle\!$ pairs to estimate the quality of generated tokens.}

\end{table*}

%% file: result.tex
\section{Evaluation Results}
We conduct experiments to study the performance of VERNet from three aspects: token-level quality estimation, sentence-level quality estimation, and the VERNet's effectiveness in GEC models. Then we present the case study to qualitatively analyze the effectiveness of the proposed two types of attention in VERNet.

\input{table/rerank.tex}
\input{figure/type_evaluation}

\subsection{Performance of Token Level Quality Estimation}
We first evaluate VERNet's effectiveness on token-level quality estimation. BERT-GED (SRC) is the previous state-of-the-art GED model~\cite{kaneko2019multi}. Additional two variants, HYP and JOINT, of BERT-GED are conducted as baselines by considering the first-ranked GEC hypothesis in beam search decoding.

As shown in Table~\ref{tab:lab_acc}, there are two scenarios, \textit{source} and \textit{hypothesis}, are conducted to evaluate model performance. The \textit{source scenario} evaluates the ability of grammaticality quality estimation, which is the same as GED models~\cite{rei2019jointly}. The \textit{hypothesis scenario} tests the quality estimation ability on GEC accuracy.

For the \textit{source scenario}, BERT-GED (JOINT) outperforms BERT-GED (SRC) and illustrates that the GEC result can help estimate the grammaticality quality of source sentences. For the \textit{hypothesis scenario}, BERT-GED (JOINT) shows better performance than BERT-GED (HYP), which thrives from the supervisions from source sentences. For \textit{both scenarios},  BERT-VERNet shows further improvement compared with BERT-GED (JOINT). Such improvements demonstrate that various GEC evidence from multiple hypotheses benefits the token-level quality estimation.

Moreover, the detection style pre-trained model ELECTRA~\cite{clark2019electra} is also used as our sentence encoder. VERNet is boosted a lot on all scenarios and datasets, which illustrates the strong ability of ELECTRA in token-level quality estimation and the generalization ability of VERNet.

\input{figure/case_att}

\subsection{Performance of Sentence Level Quality Estimation}
In this part, we evaluate VERNet's performance on sentence-level quality estimation by reranking hypotheses from beam search decoding.

Baselines can be divided into two groups: language model based and GEC accuracy based quality estimation models. The former focuses on grammaticality and fluency, including BERT-LM, BERT-GQE and BERT-GED (SRC). The others focus on estimating the GEC accuracy, including NQE, BERT-QE, BERT-GED (HYP)/(JOINT).


As shown in Table~\ref{tab:feature_rank}, we find that language model based quality estimation prefers higher recall but lower precision, which leads to more redundant corrections. Only considering grammaticality is insufficient since such unnecessary correction suggestions may mislead users. By contrast, GEC accuracy based quality estimation models get much better Precision and F$_{0.5}$, and provide more precise feedback for users. Furthermore, BERT-GED (HYP) outperforms BERT-QE, manifesting that token-level supervisions provide finer-granularity signals to help the model better distinguish subtle differences among hypotheses. VERNet outperforms all baselines, which supports our claim that multi-hypotheses from beam search provide valuable GEC evidence and help conduct more effective quality estimation for generated GEC hypotheses.


\subsection{VERNet's Effectiveness in GEC Models}
This part explores the effectiveness of VERNet on improving GEC models. We conduct VERNet$^\dagger$ by aggregating scores from the basic GEC model and VERNet for hypothesis reranking.

As shown in Table~\ref{tab:beam_rank}, two baseline models are compared in our experiments, Basic GEC~\cite{kiyono2019empirical} and BERT-fuse (GED)~\cite{kaneko2020encoder}. Compared to BERT-fuse (GED), BERT-VERNet$^\dagger$ achieves comparable performance on CoNLL-2014 and more improvement on BEA19. It demonstrates that reranking hypotheses with VERNet provides an effective way to improve basic GEC model performance without changing the Transformer architecture. R2L models incorporate four right-to-left Transformer models to improve GEC performance. However, these R2L models are not available. ELECTRA-VERNet$^\dagger$ incorporates only one model and achieves comparable performance on BEA19 and JFLEG. 

Figure~\ref{fig:type} presents VERNet$^{\dagger}$'s performance on different grammatical error types. We plot the F$_{0.5}$ scores of both basic GEC model and VERNet$^{\dagger}$ on BEA19. VERNet$^{\dagger}$ achieves improvement on most types and performs significantly better for word morphology and word usage errors, such as Noun Inflection (NOUN:INFL) and Pronoun (PRON). Such results illustrate that VERNet$^{\dagger}$ is able to leverage clues learned from multi-hypotheses to verify the GEC quality. However, we also find that VERNet$^{\dagger}$ discounts GEC performance on a few error types, \textit{e.g.}, Contraction (CONTR). The annotation biases may cause such a decrease in CONTR errors. For example, for both ``n't'' and ``not'', they are both right according to grammaticality, but annotators usually come up with different corrections with different 
GEC standards.

\subsection{Case Study}
We select one case from CoNLL-2014 and visualize node interaction and node selection attention weights to study what VERNet learns from multi-hypotheses of beam search, as shown in Figure~\ref{fig:case_attention}.

Given a source sentence, ``Do one who suffered from this disease keep it a secret of infrom their relatives ?'', and its five hypotheses from the Basic GEC Model, we plot the node interaction attention weights towards the word ``suffers'' in the hypothesis of node 2, which is assigned more higher score by BERT-VERNet. The word usage ``suffers'' is more appropriate than ``suffered'' according to the context.

The node interaction attention accurately picks up the associated tokens ``Does'' from nodes 1, 3, and 4, and ``suffers'' from node 5. ``Does'' and ``suffers'' indicate the present tense and provide sufficient evidence to verify the quality of ``suffers'' in node 2. For node selection attention, the hypothesis (node 2) shares more attention than other nodes, which is more appropriate than other hypotheses. It demonstrates that the node attention is effective to select high-quality corrections with the source-hypothesis interactions.

The attention patterns are intuitive and effective, which further demonstrates VERNet's ability to well model the interactions of multi-hypotheses for better quality estimation. 

%% file: table/rerank.tex
\begin{table*}
\small

\begin{center}
\resizebox{0.99\textwidth}{!}{
\begin{tabular}{l| p{0.55cm}<{\centering} p{0.55cm}<{\centering} p{0.55cm}<{\centering} |p{0.55cm}<{\centering} p{0.55cm}<{\centering} p{0.55cm}<{\centering} | p{0.55cm}<{\centering} p{0.55cm}<{\centering} p{0.55cm}<{\centering} | p{0.55cm}<{\centering} p{0.55cm}<{\centering} p{0.55cm}<{\centering} | p{0.7cm}<{\centering}}
\hline \multirow{2}{*}{Model} &  \multicolumn{3}{c|}{CoNLL-2014 (M$^2$)} & \multicolumn{3}{c|}{CoNLL-2014} &  \multicolumn{3}{c|}{FCE} &
\multicolumn{3}{c|}{BEA19} & JFLEG\\ \cline{2-14}
& P & R & F$_{0.5}$ & P & R & F$_{0.5}$ & P & R & F$_{0.5}$ & P & R & F$_{0.5}$ & GLEU\\ \hline
Basic GEC & 68.59 & 44.87 & 62.03 & 64.26 & 43.59 & 58.69 & 55.11 & 41.61 & 51.75 & 66.20 & \textbf{61.48} & 65.20 & 61.00  \\
Basic GEC w. R2L$^*$ & 72.4 & 46.1 & 65.0 & - & - & - & - & - & - & \textbf{74.7} & 56.7 & \textbf{70.2} & 61.4 \\
BERT-fuse (GED) & 69.2 & 45.6 & 62.6 & - & - & - & - & - & - & 67.1 & 60.1 & 65.6 & 61.3 \\
BERT-fuse (GED) w. R2L$^*$ & \textbf{72.6} & \textbf{46.4} & \textbf{65.2} & - & - & - & - & - & - & 72.3 & 61.4 & 69.8 & \textbf{62.0} \\
\hline
BERT-VERNet$^{\dagger}$ (Top2) &  69.98 & \textbf{43.69} & 62.47 & 65.62 & \textbf{41.98} & 58.98 & 58.57 & 41.53 & 54.13 & 68.42 & \textbf{60.32} & 66.63 & 61.17  \\ 
BERT-VERNet$^{\dagger}$ (Top3) & 70.49 & 43.16 & \textbf{62.57} & 65.92 & 41.22 & 58.86 & 59.20 & 41.53 & 54.55 & 69.03 & 60.20 & 67.06 & \textbf{61.24} \\ 
BERT-VERNet$^{\dagger}$ (Top4) & \textbf{70.79} & 42.72 & 62.56 & \textbf{66.65} & 40.94 & \textbf{59.21} & 59.55 & \textbf{41.55} & 54.80 & \textbf{69.43} & 60.17 & \textbf{67.36} & 61.16\\ 
BERT-VERNet$^{\dagger}$ (Top5) & 70.60 & 42.50 & 62.36 & 66.41 & 40.74 & 58.98 & \textbf{59.68} & 41.48 & \textbf{54.86} & 69.39 & 60.12 & 67.32 & 61.10 \\ 

\hline
ELECTRA-VERNet$^{\dagger}$ (Top2) & 71.21 & \textbf{44.24} & 63.47 & 66.95 & \textbf{42.97}  & 60.22 & 58.31 & 41.97 & 54.09 & 69.27 & 61.22 & 67.50 & 61.60  \\ 
ELECTRA-VERNet$^{\dagger}$ (Top3) & \textbf{71.87} & 44.13 & \textbf{63.84} & \textbf{67.51} & 42.38 & \textbf{60.35} & 59.02 & 41.99 & 54.59 & 70.64 & 61.78 & 68.67 & 61.80 \\ 
ELECTRA-VERNet$^{\dagger}$ (Top4) & 71.85 & 43.81 & 63.69 & 67.48 & 42.19 & 60.25 & 59.65 & 42.12 & 55.07 & \textbf{70.96} & \textbf{62.03} & \textbf{68.98} & \textbf{62.05}\\ 
ELECTRA-VERNet$^{\dagger}$ (Top5) & 71.58 & 43.57 & 63.43 & 67.15 & 42.10 & 60.01 & \textbf{59.95} & \textbf{42.19} & \textbf{55.29} & 70.79 & 61.74 & 68.77 & 62.07 \\ 

\hline
\end{tabular}}
\caption{\label{tab:beam_rank}Performance of Hypothesis Reranking. BERT/ELECTRA-VERNet$^{\dagger}$ aggregates the scores of Basic GEC Model~\cite{kiyono2019empirical} and VERNet for hypothesis reranking with Coordinate Ascent. BERT-fuse (GED)~\cite{kaneko2020encoder} is the Transformer model that fuses BERT representations. $^*$Note that R2L models incorporate four right-to-left Transformer models that are trained with unpublished data and these models are not supplied in their open source codes, thus these results are hard to reimplement.}
\end{center}
\end{table*}

%% file: figure/type_evaluation.tex
\begin{figure*}[t]
 \includegraphics[width=0.95\linewidth]{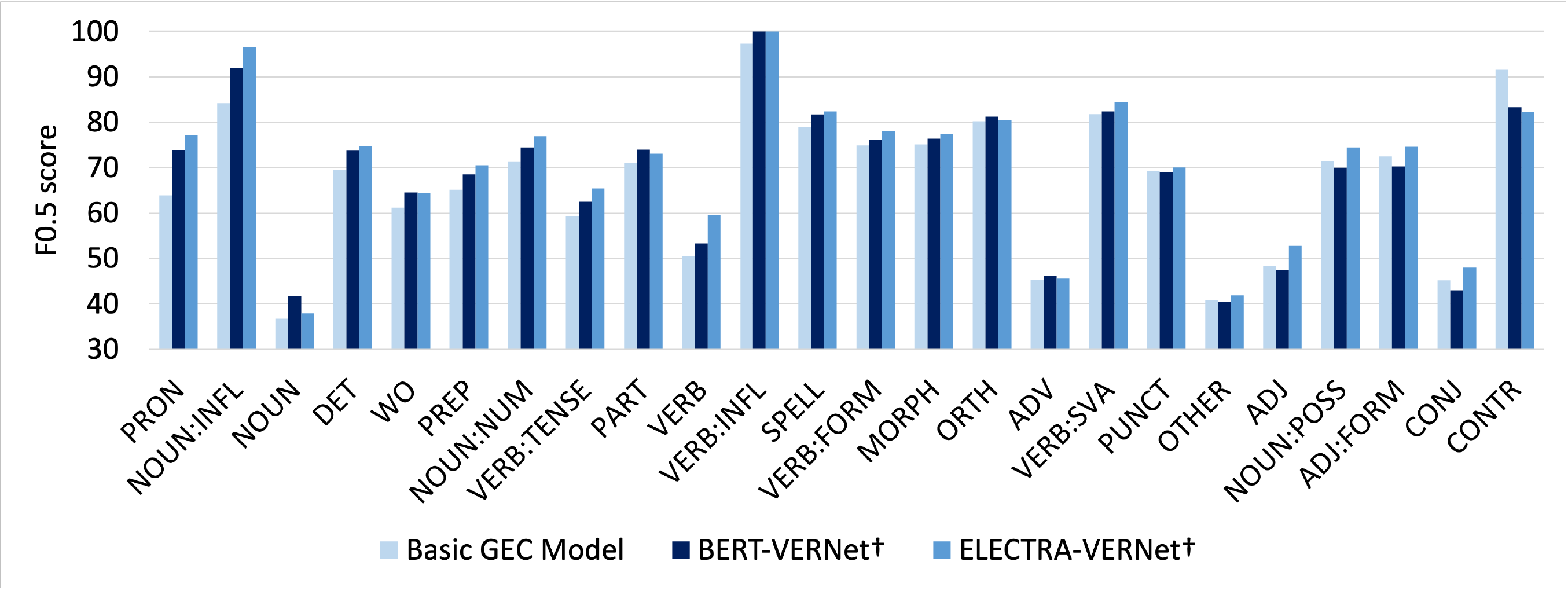}
  \caption{\label{fig:type}Model Performance of Different Grammatical Error Types on BEA19. VERNet$^{\dagger}$ reranks hypotheses with the aggregated score of basic GEC model and VERNet. All types are from ERRANT~\cite{bryant2017automatic}.}
\end{figure*}

%% file: figure/case_att.tex
\begin{figure*}[t]
\centering
 \includegraphics[width=0.90\linewidth]{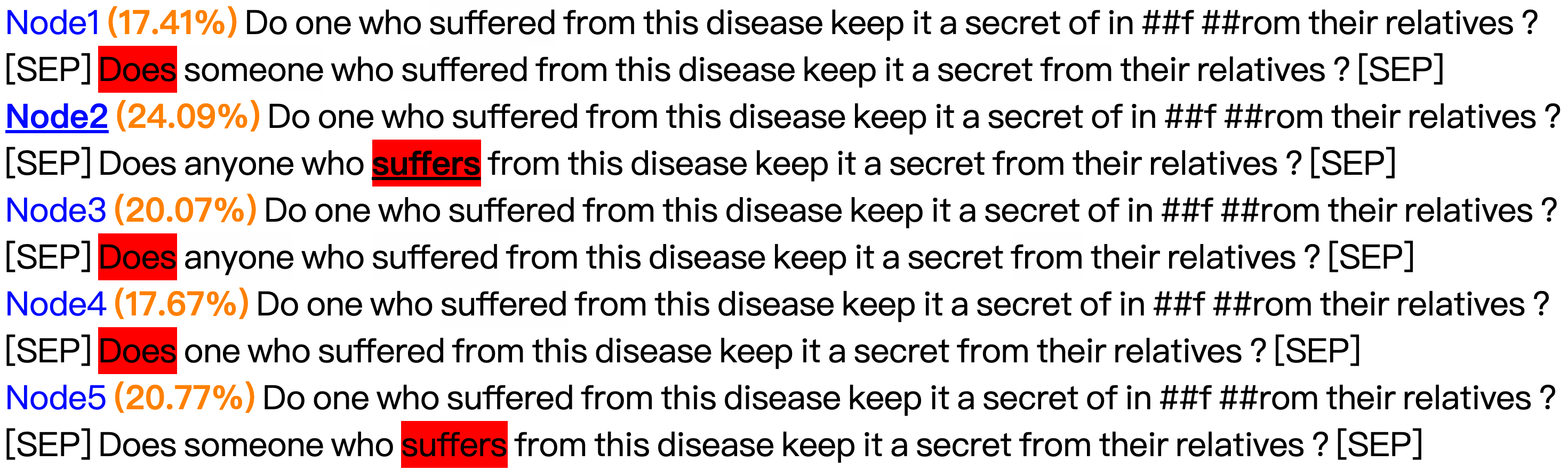}
  \caption{\label{fig:case_attention}Visualization of Attention Weight. Each node is the concatenation of the source sentence (with [SEP]) and a corresponding hypothesis sentence (with [SEP]). The selected node by BERT-VERNet is annotated (\textcolor{blue}{\textbf{{\uline{Node2}}}}). The \textcolor{orange}{node selection attention} assigned to each hypothesis is annotated with dark orange. The \textcolor{red}{node interaction attention} towards the edited token ``\textbf{\uline{suffers}}'' in the second node is also plotted. Darker red indicates higher attention weights.}
\end{figure*}

%% file: conclusion.tex
\section{Conclusion and Future Work}
This paper presents VERNet for GEC quality estimation with multi-hypotheses. VERNet models the interactions of multiple hypotheses by building a reasoning graph, and then extracts clues with two kinds of attention: \emph{node selection attention} and  \emph{node interaction attention}. They summarize and aggregate GEC evidence from multi-hypotheses to verify the quality of tokens. Experiments on four datasets show that VERNet achieves the state-of-the-art GED and quality estimation performance, and improves one published state-of-the-art GEC system. In the future, we will explore the impact of different kinds of hypotheses used in VERNet.

%% file: appendix.tex
\begin{appendix}
\section{Appendices}
\subsection{Model Details of Sentence Quality Estimation Score Calculation}\label{appendix:sentence_score}
This part describes the details of sentence score calculation of BERT based quality estimation models.

Given a source sentence $s$ with $m$ tokens and $k$-th hypothesis $c^k$ with $n$ tokens, we can get the representation $H^k$ of the $k$-th $\langle$source, hypothesis$\rangle$ sentence pair through BERT:
\begin{equation}
\small
    H^k = \text{BERT} (\text{[CLS]} \ s \ \text{[SEP]} \ c^k \ \text{[SEP]}),
\end{equation}
or only the representation $\mathcal{H}^k$ of the $k$-th hypothesis through BERT:
\begin{equation}
\small
    \mathcal{H}^k = \text{BERT} (\text{[CLS]} \ c^k \ \text{[SEP]}).
\end{equation}
The ``[CLS]'' representations are $H^k_0$ and $\mathcal{H}^k_0$.

\textbf{BERT-LM.} We mask tokens in the $k$-th hypothesis sentence $c^k$ and calculate the Perplexity of the $k$-th hypothesis sentence:
\begin{equation}
\small
    f_{\text{LM}}(c^k) = -\text{PPL}(\mathcal{H}^k_{1:n}).
\end{equation}

\textbf{BERT-GQE.} BERT-GQE uses the ``[CLS]'' representation $\mathcal{H}^k_0$ of $k$-th hypothesis to estimate the sentence quality with the probability $P(y_s|c^k)$:
\begin{equation}
\small
    P(y_s|c^k) = \text{softmax}_{y_s}(W \cdot \mathcal{H}^k_0),
\end{equation}
where $W$ is the parameter and the label $y_s$ is categorized into two groups: correct ($y_s=1$) and incorrect ($y_s=0$).

Then the sentence-level quality estimation score of hypothesis $c^k$ is calculated:
\begin{equation}
\small
    f_{\text{GQE}}(c^k) = P(y_s=1|c^k).
\end{equation}

\textbf{BERT-QE.} BERT-QE uses the ``[CLS]'' representation $H^k_0$ of $k$-th $\langle$source, hypothesis$\rangle$ sentence pair to estimate the quality of GEC hypothesis:
\begin{equation}
\small
    f_{\text{QE}}(s, c^k) = \text{sigmoid}(W \cdot H^k_0),
\end{equation}
where $W$ is the parameter. The quality estimation score $f_{\text{QE}}(s, c^k)$ of BERT-QE is trained to approximate the F$_{0.5}$ score of the $k$-th hypothesis $c^k$.

\textbf{BERT-GED.} Take BERT-GED (HYP) as an example, it uses the hypothesis representation $H^k_{m+2:m+n+2}$ of the $k$-th $\langle$source, hypothesis$\rangle$ sentence pair to estimate the quality of GEC hypothesis. Note that the ``[SEP]'' token is also used in BERT-GED to denote the end of the sentence. 

We calculate the probability of token quality estimation label $y$ for the $i$-th token $w^k_i$ in the $k$-th $\langle$source, hypothesis$\rangle$ sentence pair:
\begin{equation}
\small
    P(y|w^k_i) = \text{softmax}(W \cdot H^k_i),
\end{equation}
where $W$ is the parameter. The label $y$ is categorized into two groups: correct ($y=1$) and incorrect ($y=0$).

To estimate the quality of hypotheses, we average all token quality estimation probability $P(y=1|w^k_i)$ as the sentence quality estimation score $f(s, c^k)$ for the $k$-th  hypothesis $c^k$:
\begin{equation}
\small
    f_{\text{GED}}(s, c^k) = \frac{1}{n+1} \sum_{i=m+2}^{m+n+2} P(y=1|w^k_i).
\end{equation}

\input{table/detection0}

\subsection{Grammatical Error Detection Performance with LSTM}\label{appendix:ged}
In this experiment, we evaluate the effectiveness of BERT and LSTM on the grammatical error detection (GED) task. We keep the same setting as previous work~\cite{rei2019jointly}. The FCE dataset is used for evaluation. Precision, Recall, and F$_{0.5}$ are used as our evaluation metrics. 

As shown in Table~\ref{tab:ged}, three models, LSTM, LSTM-ATTN, and LSTM-JOINT from~\citet{rei2019jointly} are compared with the BERT model. The LSTM model leverages the LSTM encoder and adds language modeling objectives in the training process~\cite{rei2017semi}. LSTM-ATTN and LSTM-JOINT further add attention constraints and sentence level supervision to achieve better performance~\cite{rei2019jointly}. The BERT model is the same as our BERT-GED (SRC).

The BERT based model shows significant improvement than LSTM based models. Thus we do not consider LSTM based GED models in the experiments of GEC quality estimation.

\end{appendix}

%% file: table/detection0.tex
\begin{table}[t]
\begin{center}
\small
\begin{tabular}{l | c  c  c }
\hline Model & P & R & F$_{0.5}$\\
\hline
LSTM & 58.88 & 28.92 & 48.48 \\ 
BiLSTM-ATTN & 60.73 & 22.33 & 45.07 \\
BiLSTM-JOINT & 65.53 & 28.61 & 52.07 \\ \hline
BERT & \textbf{73.69} & \textbf{45.39} & \textbf{65.52} \\
\hline

\end{tabular}
\caption{\label{tab:ged}Grammatical Error Detection Performance on the First Certificate in English (FCE) dataset~\cite{yannakoudakis2011new}.
}
\end{center}
\end{table}

%% file: naacl2021.bbl
\begin{thebibliography}{44}
\expandafter\ifx\csname natexlab\endcsname\relax\def\natexlab#1{#1}\fi

\bibitem[{Awasthi et~al.(2019)Awasthi, Sarawagi, Goyal, Ghosh, and
  Piratla}]{awasthi2019parallel}
Abhijeet Awasthi, Sunita Sarawagi, Rasna Goyal, Sabyasachi Ghosh, and Vihari
  Piratla. 2019.
\newblock \href {https://www.aclweb.org/anthology/D19-1435} {Parallel iterative
  edit models for local sequence transduction}.
\newblock In \emph{Proceedings of EMNLP}, pages 4260--4270.

\bibitem[{Bryant et~al.(2019)Bryant, Felice, Andersen, and
  Briscoe}]{bryant2019bea}
Christopher Bryant, Mariano Felice, {\O}istein~E. Andersen, and Ted Briscoe.
  2019.
\newblock \href {https://www.aclweb.org/anthology/W19-4406} {The {BEA}-2019
  shared task on grammatical error correction}.
\newblock In \emph{Proceedings of the Fourteenth Workshop on Innovative Use of
  NLP for Building Educational Applications}, pages 52--75.

\bibitem[{Bryant et~al.(2017)Bryant, Felice, and Briscoe}]{bryant2017automatic}
Christopher Bryant, Mariano Felice, and Ted Briscoe. 2017.
\newblock \href {https://www.aclweb.org/anthology/P17-1074} {Automatic
  annotation and evaluation of error types for grammatical error correction}.
\newblock In \emph{Proceedings of ACL}, pages 793--805.

\bibitem[{Chollampatt and Ng(2018{\natexlab{a}})}]{chollampatt2018multilayer}
Shamil Chollampatt and Hwee~Tou Ng. 2018{\natexlab{a}}.
\newblock \href
  {https://www.aaai.org/ocs/index.php/AAAI/AAAI18/paper/view/17308} {A
  multilayer convolutional encoder-decoder neural network for grammatical error
  correction}.
\newblock In \emph{Proceedings of AAAI}, pages 5755--5762.

\bibitem[{Chollampatt and Ng(2018{\natexlab{b}})}]{chollampatt2018neural}
Shamil Chollampatt and Hwee~Tou Ng. 2018{\natexlab{b}}.
\newblock \href {https://www.aclweb.org/anthology/D18-1274} {Neural quality
  estimation of grammatical error correction}.
\newblock In \emph{Proceedings of EMNLP}, pages 2528--2539.

\bibitem[{Chollampatt et~al.(2019)Chollampatt, Wang, and
  Ng}]{chollampatt2019cross}
Shamil Chollampatt, Weiqi Wang, and Hwee~Tou Ng. 2019.
\newblock \href {https://www.aclweb.org/anthology/P19-1042} {Cross-sentence
  grammatical error correction}.
\newblock In \emph{Proceedings of ACL}, pages 435--445.

\bibitem[{Clark et~al.(2020)Clark, Luong, Le, and Manning}]{clark2019electra}
Kevin Clark, Minh{-}Thang Luong, Quoc~V. Le, and Christopher~D. Manning. 2020.
\newblock \href {https://openreview.net/forum?id=r1xMH1BtvB} {{ELECTRA:}
  pre-training text encoders as discriminators rather than generators}.
\newblock In \emph{Proceedings of ICLR}.

\bibitem[{Cui et~al.(2017)Cui, Chen, Wei, Wang, Liu, and Hu}]{cui2017attention}
Yiming Cui, Zhipeng Chen, Si~Wei, Shijin Wang, Ting Liu, and Guoping Hu. 2017.
\newblock \href {https://www.aclweb.org/anthology/P17-1055}
  {Attention-over-attention neural networks for reading comprehension}.
\newblock In \emph{Proceedings of ACL}, pages 593--602.

\bibitem[{Dahlmeier and Ng(2012)}]{dahlmeier2012better}
Daniel Dahlmeier and Hwee~Tou Ng. 2012.
\newblock \href {https://www.aclweb.org/anthology/N12-1067} {Better evaluation
  for grammatical error correction}.
\newblock In \emph{Proceedings of NAACL-HLT}, pages 568--572.

\bibitem[{Dahlmeier et~al.(2013)Dahlmeier, Ng, and Wu}]{dahlmeier2013building}
Daniel Dahlmeier, Hwee~Tou Ng, and Siew~Mei Wu. 2013.
\newblock \href {https://www.aclweb.org/anthology/W13-1703} {Building a large
  annotated corpus of learner {E}nglish: The {NUS} corpus of learner
  {E}nglish}.
\newblock In \emph{Proceedings of the Eighth Workshop on Innovative Use of
  {NLP} for Building Educational Applications}, pages 22--31.

\bibitem[{Devlin et~al.(2019)Devlin, Chang, Lee, and
  Toutanova}]{devlin2019bert}
Jacob Devlin, Ming-Wei Chang, Kenton Lee, and Kristina Toutanova. 2019.
\newblock \href {https://www.aclweb.org/anthology/N19-1423} {{BERT}:
  Pre-training of deep bidirectional transformers for language understanding}.
\newblock In \emph{Proceedings of NAACL-HLT}, pages 4171--4186.

\bibitem[{Felice et~al.(2016)Felice, Bryant, and Briscoe}]{felice2016automatic}
Mariano Felice, Christopher Bryant, and Ted Briscoe. 2016.
\newblock \href {https://www.aclweb.org/anthology/C16-1079} {Automatic
  extraction of learner errors in {ESL} sentences using linguistically enhanced
  alignments}.
\newblock In \emph{Proceedings of COLING}, pages 825--835.

\bibitem[{Fomicheva et~al.(2020)Fomicheva, Specia, and
  Guzm{\'a}n}]{fomicheva2020multi}
Marina Fomicheva, Lucia Specia, and Francisco Guzm{\'a}n. 2020.
\newblock \href {https://www.aclweb.org/anthology/2020.acl-main.113}
  {Multi-hypothesis machine translation evaluation}.
\newblock In \emph{Proceedings of ACL}, pages 1218--1232.

\bibitem[{Ge et~al.(2018)Ge, Wei, and Zhou}]{ge2018fluency}
Tao Ge, Furu Wei, and Ming Zhou. 2018.
\newblock \href {https://www.aclweb.org/anthology/P18-1097} {Fluency boost
  learning and inference for neural grammatical error correction}.
\newblock In \emph{Proceedings of ACL}, pages 1055--1065.

\bibitem[{Grundkiewicz et~al.(2019)Grundkiewicz, Junczys-Dowmunt, and
  Heafield}]{grundkiewicz2019neural}
Roman Grundkiewicz, Marcin Junczys-Dowmunt, and Kenneth Heafield. 2019.
\newblock \href {https://www.aclweb.org/anthology/W19-4427} {Neural grammatical
  error correction systems with unsupervised pre-training on synthetic data}.
\newblock In \emph{Proceedings of the Fourteenth Workshop on Innovative Use of
  NLP for Building Educational Applications}, pages 252--263.

\bibitem[{Hagiwara and Mita(2020)}]{hagiwara2019github}
Masato Hagiwara and Masato Mita. 2020.
\newblock \href {https://www.aclweb.org/anthology/2020.lrec-1.835} {{G}it{H}ub
  typo corpus: A large-scale multilingual dataset of misspellings and
  grammatical errors}.
\newblock In \emph{Proceedings of the 12th Language Resources and Evaluation
  Conference}, pages 6761--6768.

\bibitem[{Hoang et~al.(2016)Hoang, Chollampatt, and Ng}]{hoang2016exploiting}
Duc~Tam Hoang, Shamil Chollampatt, and Hwee~Tou Ng. 2016.
\newblock \href {http://www.ijcai.org/Abstract/16/398} {Exploiting n-best
  hypotheses to improve an {SMT} approach to grammatical error correction}.
\newblock In \emph{Proceedings of IJCAI}, pages 2803--2809.

\bibitem[{Junczys-Dowmunt et~al.(2018)Junczys-Dowmunt, Grundkiewicz, Guha, and
  Heafield}]{junczys2018approaching}
Marcin Junczys-Dowmunt, Roman Grundkiewicz, Shubha Guha, and Kenneth Heafield.
  2018.
\newblock \href {https://www.aclweb.org/anthology/N18-1055} {Approaching neural
  grammatical error correction as a low-resource machine translation task}.
\newblock In \emph{Proceedings of NAACL-HLT}, pages 595--606.

\bibitem[{Kaneko et~al.(2019)Kaneko, Hotate, Katsumata, and
  Komachi}]{kaneko2019tmu}
Masahiro Kaneko, Kengo Hotate, Satoru Katsumata, and Mamoru Komachi. 2019.
\newblock \href {https://www.aclweb.org/anthology/W19-4422} {{TMU} transformer
  system using {BERT} for re-ranking at {BEA} 2019 grammatical error correction
  on restricted track}.
\newblock In \emph{Proceedings of the Fourteenth Workshop on Innovative Use of
  NLP for Building Educational Applications}, pages 207--212.

\bibitem[{Kaneko and Komachi(2019)}]{kaneko2019multi}
Masahiro Kaneko and Mamoru Komachi. 2019.
\newblock \href
  {https://www.cys.cic.ipn.mx/ojs/index.php/CyS/article/view/3271} {Multi-head
  multi-layer attention to deep language representations for grammatical error
  detection}.
\newblock \emph{Computaci{\'o}n y Sistemas}, (3).

\bibitem[{Kaneko et~al.(2020)Kaneko, Mita, Kiyono, Suzuki, and
  Inui}]{kaneko2020encoder}
Masahiro Kaneko, Masato Mita, Shun Kiyono, Jun Suzuki, and Kentaro Inui. 2020.
\newblock \href {https://www.aclweb.org/anthology/2020.acl-main.391}
  {Encoder-decoder models can benefit from pre-trained masked language models
  in grammatical error correction}.
\newblock In \emph{Proceedings of ACL}, pages 4248--4254.

\bibitem[{Kingma and Ba(2015)}]{kingma2014adam}
Diederik~P. Kingma and Jimmy Ba. 2015.
\newblock \href {https://openreview.net/forum?id=8gmWwjFyLj} {Adam: {A} method
  for stochastic optimization}.
\newblock In \emph{Proceedings of ICLR}.

\bibitem[{Kiyono et~al.(2019)Kiyono, Suzuki, Mita, Mizumoto, and
  Inui}]{kiyono2019empirical}
Shun Kiyono, Jun Suzuki, Masato Mita, Tomoya Mizumoto, and Kentaro Inui. 2019.
\newblock \href {https://www.aclweb.org/anthology/D19-1119} {An empirical study
  of incorporating pseudo data into grammatical error correction}.
\newblock In \emph{Proceedings of EMNLP}, pages 1236--1242.

\bibitem[{Lichtarge et~al.(2019)Lichtarge, Alberti, Kumar, Shazeer, Parmar, and
  Tong}]{lichtarge2019corpora}
Jared Lichtarge, Chris Alberti, Shankar Kumar, Noam Shazeer, Niki Parmar, and
  Simon Tong. 2019.
\newblock \href {https://www.aclweb.org/anthology/N19-1333} {Corpora generation
  for grammatical error correction}.
\newblock In \emph{Proceedings of NAACL-HLT}, pages 3291--3301.

\bibitem[{Malmi et~al.(2019)Malmi, Krause, Rothe, Mirylenka, and
  Severyn}]{malmi2019encode}
Eric Malmi, Sebastian Krause, Sascha Rothe, Daniil Mirylenka, and Aliaksei
  Severyn. 2019.
\newblock \href {https://www.aclweb.org/anthology/D19-1510} {Encode, tag,
  realize: High-precision text editing}.
\newblock In \emph{Proceedings of EMNLP}, pages 5054--5065.

\bibitem[{Metzler and Croft(2007)}]{metzler2007linear}
Donald Metzler and W~Bruce Croft. 2007.
\newblock \href
  {https://link.springer.com/content/pdf/10.1007/s10791-006-9019-z.pdf} {Linear
  feature-based models for information retrieval}.
\newblock \emph{Information Retrieval}.

\bibitem[{Mizumoto et~al.(2011)Mizumoto, Komachi, Nagata, and
  Matsumoto}]{Mizumoto2011MiningRL}
Tomoya Mizumoto, Mamoru Komachi, Masaaki Nagata, and Yuji Matsumoto. 2011.
\newblock \href {https://www.aclweb.org/anthology/I11-1017} {Mining revision
  log of language learning {SNS} for automated {J}apanese error correction of
  second language learners}.
\newblock In \emph{Proceedings of IJCNLP}, pages 147--155.

\bibitem[{Napoles et~al.(2015)Napoles, Sakaguchi, Post, and
  Tetreault}]{napoles2015ground}
Courtney Napoles, Keisuke Sakaguchi, Matt Post, and Joel Tetreault. 2015.
\newblock \href {https://www.aclweb.org/anthology/P15-2097} {Ground truth for
  grammatical error correction metrics}.
\newblock In \emph{Proceedings of ACL}, pages 588--593.

\bibitem[{Napoles et~al.(2017)Napoles, Sakaguchi, and
  Tetreault}]{napoles2017jfleg}
Courtney Napoles, Keisuke Sakaguchi, and Joel Tetreault. 2017.
\newblock \href {https://www.aclweb.org/anthology/E17-2037} {{JFLEG}: A fluency
  corpus and benchmark for grammatical error correction}.
\newblock In \emph{Proceedings of EACL}, pages 229--234.

\bibitem[{Ng et~al.(2014)Ng, Wu, Briscoe, Hadiwinoto, Susanto, and
  Bryant}]{ng2014conll}
Hwee~Tou Ng, Siew~Mei Wu, Ted Briscoe, Christian Hadiwinoto, Raymond~Hendy
  Susanto, and Christopher Bryant. 2014.
\newblock \href {https://www.aclweb.org/anthology/W14-1701} {The {C}o{NLL}-2014
  shared task on grammatical error correction}.
\newblock In \emph{Proceedings of the Eighteenth Conference on Computational
  Natural Language Learning: Shared Task}, pages 1--14.

\bibitem[{Omelianchuk et~al.(2020)Omelianchuk, Atrasevych, Chernodub, and
  Skurzhanskyi}]{omelianchuk2020gector}
Kostiantyn Omelianchuk, Vitaliy Atrasevych, Artem Chernodub, and Oleksandr
  Skurzhanskyi. 2020.
\newblock \href {https://www.aclweb.org/anthology/2020.bea-1.16} {{GECT}o{R}
  {--} grammatical error correction: Tag, not rewrite}.
\newblock In \emph{Proceedings of the Fifteenth Workshop on Innovative Use of
  NLP for Building Educational Applications}, pages 163--170.

\bibitem[{Ott et~al.(2018)Ott, Auli, Grangier, and Ranzato}]{ott2018analyzing}
Myle Ott, Michael Auli, David Grangier, and Marc'Aurelio Ranzato. 2018.
\newblock \href {http://proceedings.mlr.press/v80/ott18a.html} {Analyzing
  uncertainty in neural machine translation}.
\newblock In \emph{Proceedings of ICML}, pages 3953--3962.

\bibitem[{Rei(2017)}]{rei2017semi}
Marek Rei. 2017.
\newblock \href {https://www.aclweb.org/anthology/P17-1194} {Semi-supervised
  multitask learning for sequence labeling}.
\newblock In \emph{Proceedings of ACL}, pages 2121--2130.

\bibitem[{Rei and S{\o}gaard(2019)}]{rei2019jointly}
Marek Rei and Anders S{\o}gaard. 2019.
\newblock \href {https://doi.org/10.1609/aaai.v33i01.33016916} {Jointly
  learning to label sentences and tokens}.
\newblock In \emph{Proceedings of AAAI}, pages 6916--6923.

\bibitem[{Sutskever et~al.(2014)Sutskever, Vinyals, and
  Le}]{sutskever2014sequence}
Ilya Sutskever, Oriol Vinyals, and Quoc~V. Le. 2014.
\newblock \href
  {https://proceedings.neurips.cc/paper/2014/hash/a14ac55a4f27472c5d894ec1c3c743d2-Abstract.html}
  {Sequence to sequence learning with neural networks}.
\newblock In \emph{Proceedings of NIPS}, pages 3104--3112.

\bibitem[{Vaswani et~al.(2017)Vaswani, Shazeer, Parmar, Uszkoreit, Jones,
  Gomez, Kaiser, and Polosukhin}]{Vaswani2017AttentionIA}
Ashish Vaswani, Noam Shazeer, Niki Parmar, Jakob Uszkoreit, Llion Jones,
  Aidan~N. Gomez, Lukasz Kaiser, and Illia Polosukhin. 2017.
\newblock \href
  {https://proceedings.neurips.cc/paper/2017/hash/3f5ee243547dee91fbd053c1c4a845aa-Abstract.html}
  {Attention is all you need}.
\newblock In \emph{Proceedings of NIPS}, pages 5998--6008.

\bibitem[{Wang et~al.(2019{\natexlab{a}})Wang, Zhao, Jia, Li, and
  Liu}]{wang2019denoising}
Liang Wang, Wei Zhao, Ruoyu Jia, Sujian Li, and Jingming Liu.
  2019{\natexlab{a}}.
\newblock \href {https://www.aclweb.org/anthology/D19-1412} {Denoising based
  sequence-to-sequence pre-training for text generation}.
\newblock In \emph{Proceedings of EMNLP}, pages 4003--4015.

\bibitem[{Wang et~al.(2019{\natexlab{b}})Wang, Liu, Wang, Luan, and
  Sun}]{wang2019improving}
Shuo Wang, Yang Liu, Chao Wang, Huanbo Luan, and Maosong Sun.
  2019{\natexlab{b}}.
\newblock \href {https://www.aclweb.org/anthology/D19-1073} {Improving
  back-translation with uncertainty-based confidence estimation}.
\newblock In \emph{Proceedings of EMNLP}, pages 791--802.

\bibitem[{Wolf et~al.(2020)Wolf, Debut, Sanh, Chaumond, Delangue, Moi, Cistac,
  Rault, Louf, Funtowicz, Davison, Shleifer, von Platen, Ma, Jernite, Plu, Xu,
  Le~Scao, Gugger, Drame, Lhoest, and Rush}]{wolf2020transformers}
Thomas Wolf, Lysandre Debut, Victor Sanh, Julien Chaumond, Clement Delangue,
  Anthony Moi, Pierric Cistac, Tim Rault, Remi Louf, Morgan Funtowicz, Joe
  Davison, Sam Shleifer, Patrick von Platen, Clara Ma, Yacine Jernite, Julien
  Plu, Canwen Xu, Teven Le~Scao, Sylvain Gugger, Mariama Drame, Quentin Lhoest,
  and Alexander Rush. 2020.
\newblock \href {https://www.aclweb.org/anthology/2020.emnlp-demos.6/}
  {Transformers: State-of-the-art natural language processing}.
\newblock In \emph{Proceedings of EMNLP}, pages 38--45.

\bibitem[{Xie et~al.(2018)Xie, Genthial, Xie, Ng, and
  Jurafsky}]{xie2018noising}
Ziang Xie, Guillaume Genthial, Stanley Xie, Andrew Ng, and Dan Jurafsky. 2018.
\newblock \href {https://www.aclweb.org/anthology/N18-1057} {Noising and
  denoising natural language: Diverse backtranslation for grammar correction}.
\newblock In \emph{Proceedings of NAACL-HLT}, pages 619--628.

\bibitem[{Yannakoudakis et~al.(2011)Yannakoudakis, Briscoe, and
  Medlock}]{yannakoudakis2011new}
Helen Yannakoudakis, Ted Briscoe, and Ben Medlock. 2011.
\newblock \href {https://www.aclweb.org/anthology/P11-1019} {A new dataset and
  method for automatically grading {ESOL} texts}.
\newblock In \emph{Proceedings of ACL}, pages 180--189.

\bibitem[{Yannakoudakis et~al.(2017)Yannakoudakis, Rei, Andersen, and
  Yuan}]{yannakoudakis2017neural}
Helen Yannakoudakis, Marek Rei, {\O}istein~E. Andersen, and Zheng Yuan. 2017.
\newblock \href {https://www.aclweb.org/anthology/D17-1297} {Neural
  sequence-labelling models for grammatical error correction}.
\newblock In \emph{Proceedings of EMNLP}, pages 2795--2806.

\bibitem[{Yuan and Briscoe(2016)}]{yuan2016grammatical}
Zheng Yuan and Ted Briscoe. 2016.
\newblock \href {https://www.aclweb.org/anthology/N16-1042} {Grammatical error
  correction using neural machine translation}.
\newblock In \emph{Proceedings of NAACL-HLT}, pages 380--386.

\bibitem[{Zhao et~al.(2019)Zhao, Wang, Shen, Jia, and Liu}]{zhao2019improving}
Wei Zhao, Liang Wang, Kewei Shen, Ruoyu Jia, and Jingming Liu. 2019.
\newblock \href {https://www.aclweb.org/anthology/N19-1014} {Improving
  grammatical error correction via pre-training a copy-augmented architecture
  with unlabeled data}.
\newblock In \emph{Proceedings of NAACL-HLT}, pages 156--165.

\end{thebibliography}
